\documentclass[sigconf]{acmart}

\usepackage[ruled,vlined,linesnumbered]{algorithm2e}
\usepackage{multirow}
\usepackage{url}
\renewcommand\footnotetextcopyrightpermission[1]{}

\begin{document}

\title{Path-aware Siamese Graph Neural Network for Link Prediction}

\author{Jingsong Lv, Zhao Li, Hongyang Chen, Yao Qi, Chunqi Wu}
\affiliation{%
  \institution{Research Center for Graph Computing, Zhejiang Lab}
  \city{Hangzhou}
  \state{Zhejiang Province}
  \country{China}
}
\email{{jingsonglv,hongyang,qiy}@zhejianglab.com, lzjoey@gmail.com, francis_chun@163.com}

\renewcommand{\shortauthors}{Jingsong Lv, et al.}

\begin{abstract}
  In this paper, we propose an algorithm of Path-aware Siamese Graph neural network(PSG) for link prediction tasks. First, PSG captures both nodes and edge features for given two nodes, namely the structure information of k-neighborhoods and relay paths information of the nodes. Furthermore, a novel multi-task GNN framework with self-supervised contrastive learning is proposed for differentiation of positive links and negative links while content and behavior of nodes can be captured simultaneously. We evaluate the proposed algorithm PSG  on two link property prediction datasets, ogbl-ddi and ogbl-collab. PSG achieves top 1 performance on ogbl-ddi until submission and top 3 performance on ogbl-collab. The experimental results verify the superiority of our proposed PSG.
\end{abstract}



\maketitle

\section{Introduction}
The task of link prediction is often used to predict missing links in static networks.  It is widely applied in the scenarios of recommender system~\cite{bennett2007netflix}, social network analysis~\cite{adamic2003friends}, bio-computing~\cite{zhao2018bipartite}, computational pharmacy~\cite{stanfield2017drug}, knowledge graph~\cite{nickel2015review} and
 other graph based network analysis. Generally speaking, it can be inferenced by computing the similarity of the  nodes to judge whether missing links exist between two nodes. This straightforward idea leads to research direction on the similarity measures and representation learing of graph nodes. Intuitively, common neighbors can measure the similarity of two nodes, therefore stimulate heuristic based methods on the context of nodes. Furthermore, network embedding methods are studied  as representation learning to explore latent factors as feature vectors of nodes. Recently, Graph Neural Networks (GNNs) based methods have been widely applied to the work of neural link predition due to its nonlinear feature extraction power of graph structure~\cite{wu2020comprehensive,zhou2020graph}.

\par Given ogbl-ddi dataset of Open Graph Benchmark (OGB)~\cite{hu2020ogb}, which is a homogeneous, unweighted, undirected graph, representing the drug-drug interaction network ~\cite{wishart2018drugbank}, the objective is to rank true drug-drug interaction higher than false drug-drug interaction. In the OGB leaderboard, existing approaches  mainly focus on GNNs that capture node features of neighborhood, while some ones integrate edge features, such as distance encoding~\cite{li2020distance}, into existing GNN framework. In addition, based on GraphSage~\cite{hamilton2017inductive}, PLNLP ~\cite{wang2021pairwise} introduces a pairwise loss funtion, which is common in siamese network for ranking problem, and achieves better performance.  

In this paper, we propose a Path-aware Siamese Graph neural network(PSG) for link prediction tasks. Our contributions can be summarized as follows: (1) \textbf{A Novel GNN-Friendly Algorithm.} We propose the first Path-aware Siamese GNN algorithm for link prediction, which can be easily adapted to existing backbone GNN networks, such as GraphSage, PLNLP, and GIDN~\cite{wang2022gidn}. (2) \textbf{A Novel Self-Supervised Representation Learning Framework.} We introduce clustering to get node soft labels and propose a novel multi-task representation learning framework to simultaneously learn label task and link prediction task. Therefore, positive links and negative links can be differentiated by pairwise learning while both content and behavior of nodes are aligned by contrastive learning. (3) \textbf{High Performance.} We evaluate the proposed algorithm PSG  on two link property prediction datasets of Open Graph Benchmark (OGB), ogbl-ddi and ogbl-collab. The experimental results demonstrate the superiority of the PSG.

In the following, we present our work in four sections. Section ~\Ref{section_relatedwork} introduces related work on link prediction and GNN. Section ~\Ref{section_method} formally defines the link prediction problem, and presents our solution PSG model. Section ~\Ref{section_exp} shows experiments and evaluation details of the model. Finally, Section ~\Ref{section_conclusion} summarizes this work.

\section{Related Work} \label{section_relatedwork}
\subsection{Link Prediction} For the task of link prediction,  approaches can be summarized as follows: neighborhood based, factor decomposition based and neural network based. Neighborhood based methods mainly utilize wide-based or depth-based neighborhood information to measure nodes similarity, such as common neighbors~\cite{liben2007link},  Adamic-Adar~\cite{adamic2003friends}, SimRank~\cite{jeh2002simrank} and Node2Vec~\cite{grover2016node2vec}. As the former method focus only on the explicit factors of graph, factor decomposition based method PNRL~\cite{wang2017predictive}. Furthermore, to capture deep hierarchical structure features, neural network based methods are proposed to perform end-to-end encoding of graph structure features with neural layers, such as GCN~\cite{kipf2016semi}, SEAL~\cite{zhang2018link}. 

\subsection{Graph Neural Network} Deep neural networks have achieved great success in Euclidean domains, such as computer vision~\cite{liu2021swin}, speech synthesis and recognition~\cite{snyder2018x}, natural language understanding~\cite{devlin2018bert}. It inspires the research of applying GNNs on non-Euclidean domains. In earlier stage, based on the theory of convolution in digital signal process, spectral domain-based GNNs attract lots of attention and research~\cite{wu2020comprehensive,zhou2020graph}. Graph Convolution Network(GCN)~\cite{kipf2016semi} is one of the most famous models that achieves better performance than traditional methods. As time and space complexity of GCN is high,  variants of spectral domain based methods~\cite{defferrard2016convolutional, levie2018cayleynets, huang2022graph} aim to reduce the time cost and space cost, and to generate better approximations and performance. Furthermore, spatial domain based methods are proposed to directly encode local structure features based on neighborhood aggregation and graph pooling~\cite{gilmer2017neural}. As computations increase exponentially with the increase of neighborhood range, sampling ~\cite{hamilton2017inductive} and attentions methods~\cite{velivckovic2017graph} are proposed to solve the problem of computation efficiency. As a side effect, the methods often improve the final performance as the extracted information and the capacity of neural layers increase. These methods can also be regarded as graph re-connection approaches. To further improve the effectiveness and efficiency, few-shot learning~\cite{lin2022structure}, contrastive learning~\cite{liu2021anomaly} are also introduced and integrated into GNNs.

\section{Method}  \label{section_method}
In this section, we present the problem definition and our solution PSG model.

\subsection{Problem Formulation}
Let $G=(V,E)$ represent a homogeneous, unweighted, undirected graph, where $V$ represents the nodes set of the graph, and $E \subseteq V \times V$ represents the edges set of the graph. Let $N=|V|$ and $M=|E|$. Given an edge $e_{i,j} \in E$, it can be represented as node pair $(v_i, v_j)$, where $v_i,v_j \in V$.  Now let $A \in \mathbb{R}^{N \times N}$ represent the corresponding symmetric adjacency matrix, where matrix element $a_{i,j}$ indicates whether edge $(v_i, v_j) \in E$. For node features, let $X\in \mathbb{R}^{N \times d}$ represent the feature matrix, where the matrix element $x_{i,j}$ represents the $j_{th}$ element of feature vector of node $v_i$.

Now given a  network $G=(V,E)$, the problem is to predict whether a missing link edge $e' \in E'=  V \times V - E$ exists. 
Suppose the prediction value of a positive sample is higher than that of a negative sample, namely $1 \geq f(e^+ | \theta) > f(e^- | \theta) \geq 0$, where $f(\cdot)$ is prediction model function,   $e^+$ is a positive edge sample and $e^-$ is a uncertain negative sample. Intuitively, a harder learning objective is to rank all edge samples by maximizing the difference of $f(e^+ | \theta)$ and $f(e^- | \theta)$. Formally, we optimize the model by the following objective loss function, squared least surrogate loss~\cite{gao2015consistency, wang2021pairwise}:

\begin{equation}
L(\theta) = \min_{\theta} \sum\limits_{e^+ \in E, e^- \in E'}{(1-f(e^+| \theta) + f(e^-| \theta))^2 + \frac{\lambda}{2}\Vert\theta\Vert^2} \text{,}
\label{model_loss}
\end{equation}
\noindent where $\lambda$ is a weight hyper-parameter  of L2 regularization. 

\subsection{Model Buildup}
To solve the problem described in the former subsection, we propose a Path-aware Siamese GNN model, or simply PSG model. In the following, we introduce the details of the  model.

\subsubsection{Model Architecture}
\begin{figure*}[htbp]
\includegraphics[width=100mm,scale=0.5]{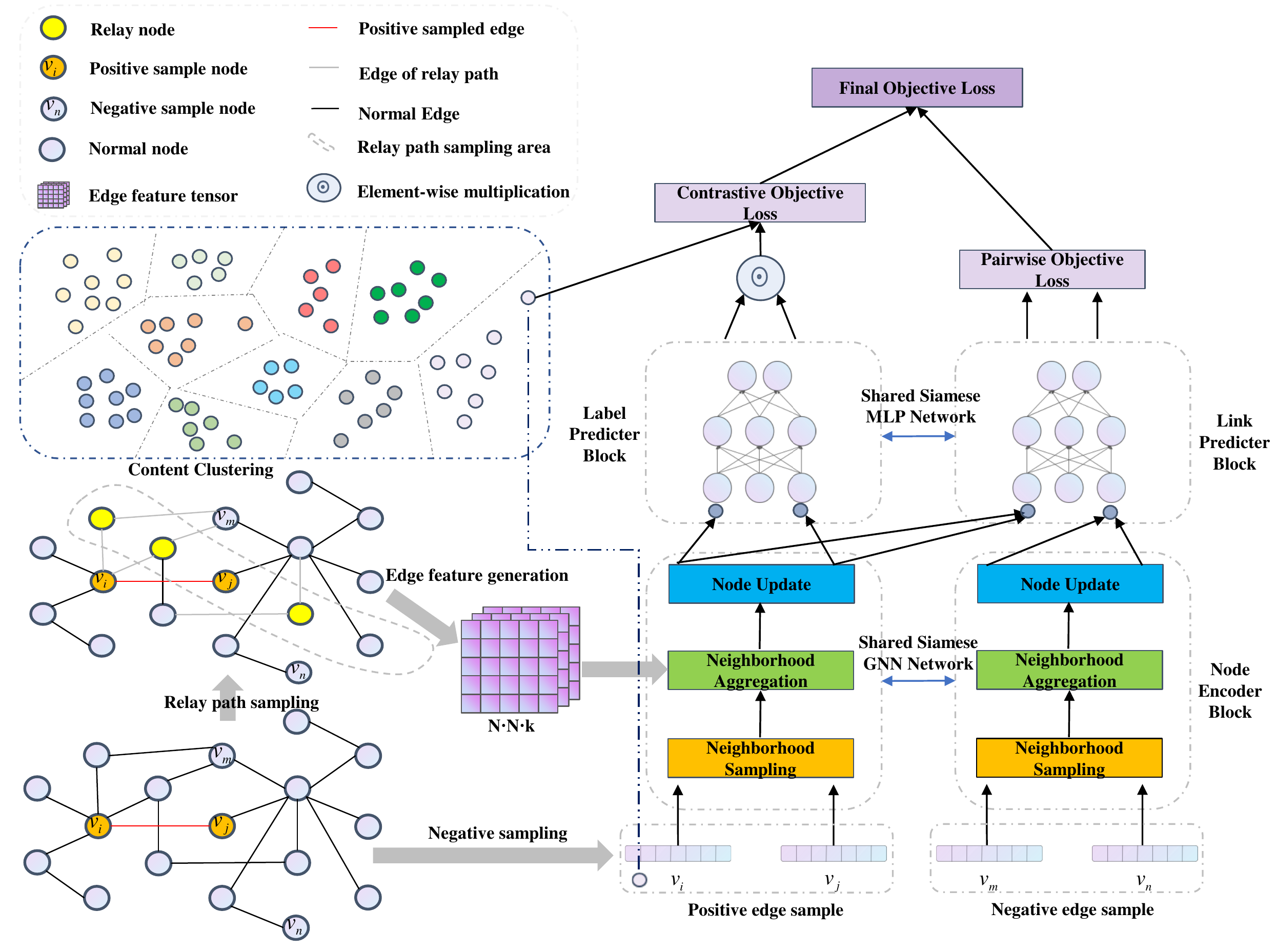}
\caption{Overview of PSG model architecture.} \label{fig_model}
\centering
\end{figure*}

In general, PSG model is a combinatorial model, which can be represented as a function of combination of two associated paired nodes, as shown in Formula ~\Ref{model_comb}.
\begin{equation}
f(e_{i,j}| \theta)  = f(v_i, v_j| \theta) \label{model_comb} \text{.}
\end{equation}

Each associated node is learned and represented as a hidden feature embedding based on a representation learning function. So the model can be reformed as a link prediction function of representation function of two associated nodes, as shown in Formula ~\Ref{model_gh}.
\begin{equation}
f(v_i, v_j| \theta) = g(h_1(v_i | \theta^{h_1}),  h_2(v_j | \theta^{h_2}) | \theta^g) \label{model_gh} \text{.}
\end{equation}

For undirected graph, representation functions of nodes can share the same one, as shown in Formula ~\Ref{model_emb}.
\begin{equation}
h_1(v | \theta^{h_1}) =  h_2(v | \theta^{h_2}) \triangleq h(v | \theta^{h}) \label{model_emb} \text{.}
\end{equation}

Here, GNN network is adopted as a representation function model of graph nodes, as shown in Formula ~\Ref{model_gnn}.

\begin{equation}
h(v | \theta^{h}) = GNN(v | \theta^{h} ) \label{model_gnn} \text{.}
\end{equation}

And multi-layer perceptron (MLP) is applied as link prediction function, as shown in Formula  ~\Ref{model_mlp}.
\begin{equation}
g(h_1, h_2| \theta^g) = MLP(h_1, h_2 | \theta^g) \label{model_mlp} \text{.}
\end{equation}

Based on the above combinatorial model framework, PSG can be divided into four parts: edge featuring, node encoder, link predictor and pairwise objective loss function. The overview of the PSG model architecture is shown in Figure ~\Ref{fig_model}.

In the following, we present more details of the four parts.

\subsubsection{Edge Featuring} For a traditional machine learning task, feature engineering can directly improve the model performance due to correlation improvement of features.  Therefore, a feasible solution is to utilize strong edge feature. It can be implemented by calculating the Shortest Path Distance(SPD)~\cite{li2020distance, shitao2021struct} for any given paired nodes, or simply called edge. To improve the path structure capturing ability, a relay path sampling method is applied to generate multiple relay path-aware SPDs. For robustness, each relay path sampling selects a random sampling area and relay nodes. Then, a relay path-aware edge feature is generated as expectation of SPDs through relay paths. Such a process is executed for several times, namely $k$. Hence, we get edge feature vector $h_{uv}$ with dimension $k$. Finally, an edge feature tensor with dimension $N \times N \times k$ is generated.
\begin{equation}
\begin{split}
 \mathbf{h}_{uv} =  CONCAT_{k}[Avg_{r \in RelayArea}(SPD_{u\mathbf{r}} + SPD_{\mathbf{r}v})] \text{,}
 \label{model_relaysample} 
\end{split}
\end{equation}
\noindent where $CONCAT_{k}$ means concatenation $k$ times, $Avg$ represents Average operator, $r \in RelayArea$ represents a relay node in relay sampling area, and $SPD_{u\mathbf{r}}$ represents the Shortest Path Distance between nodes $u$ and $r$.

\subsubsection{Node Encoder} For representation of two associated nodes, a shared siamese GNN model is used to encode the node features and path-aware edge features of neighborhood. The model shares the network structure and weights for representation of associated nodes from positive and negative edge samples. Similar to GraphSAGE~\cite{hamilton2017inductive}, the node encoder block mainly contains neighborhood sampling, neighborhood aggregation and node update. Besides, path-aware edge features are integrated into the shared siamese GNN network, which improves the contrastive learning of different relay path features between positive  and negative edge samples. Specifically, Formula ~\Ref{model_gnn} can  be depicted by Formula ~\Ref{model_sageedge}.

\begin{equation}
\begin{split}
 GNN(v | \theta^{h} ) \sim \mathbf{h}_{v}^{l+1}=Relu(\mathbf{W_1}^{l+1} \mathbf{h}_v^l + \\
 \mathbf{W_2}^{l+1}  \mathcal{A}gg_{u \in N^r(v)}(Relu(\mathbf{h}_u^l + \mathbf{W_3} \mathbf{h}_{uv})))  \label{model_sageedge} \text{,}
\end{split}
\end{equation}
\noindent where $N^r(v)$ represents the $r$-hop neighborhood node set of node $v$, in which the distance to node $v$ is not greater than $r$. And $ \mathcal{A}gg$ represents the aggregation function of neighborhood nodes, which is set to $Mean$ as default.

\subsubsection{Link Predictor} As a decoder, MLP~\cite{murtagh1991multilayer} is adopted as link predictor function. Like node encoder block, link predictor block also shares the network structure and weights for decoding associated nodes. In addition, feature intersection is applied within this block. Formula ~\Ref{model_mlp} is rewritten as Formula ~\Ref{model_mlpreform}.
\begin{equation}
MLP(h_1, h_2 | \theta^g) = Relu(\mathbf{W_4}(h_1 \circ h_2)) \label{model_mlpreform} \text{,}
\end{equation}
\noindent where $\circ$ represents Hadamard product and $\mathbf{W_4}$ is the corresponding weight matrix.

\subsubsection{Pairwise Objective Loss Function } To guarantee the generalized performance of the model, the former pairwise objective loss function depicted by Formula ~\Ref{model_loss} is reformed as Formula ~\Ref{model_finalloss}. 

\begin{equation}
L_h(\theta) = \min_{\theta} \sum\limits_{e^+ \in E, e^- \in E'}{(1-f(e^+| \theta) + f(e^-| \theta))^2 + \frac{1}{2}\sum\limits_{i=1}^{4}{\lambda_i\Vert\mathbf{W_i}\Vert^2}} \label{model_finalloss} \text{,}
\end{equation}
\noindent where $\mathbf{W_i} \in \theta$ is a weight matrix of the parameter $\theta$, and $\lambda_i$ is the corresponding weight hyper-parameter.

\subsection{Contrastive Learning}
Based on the former backbone GNN network, node embeddings can be learned and generated as representations of nodes interaction behaviors, called behavior representations. In the condition of unsupervised settings, the distribution of behavior representations can be different from the distribution of nodes content representations. To alleviate the deviation between the distributions of behavior representations and content representations, we introduce node content clustering, label predictor and multi-task learning as a guide of the former backbone GNN network, or simply contrastive learning of behavior representations and content representations.

\subsubsection{Content Clustering}
For clustering of high dimensional vectors, a common approach is K-means~\cite{sculley2010web} due to its efficiency. We adopt it to generate the content labels of nodes as Formula \Ref{clustering_kmeans}.
\begin{equation}
Cl(v_{i}| \theta^{Cl})  \triangleq c_i = Kmeans(v_i| \theta^{Cl}) \label{clustering_kmeans}  \text{.}
\end{equation}

\subsubsection{Label Predictor}
Like link predictor, MLP~\cite{murtagh1991multilayer} is also adopted as label predictor function. And the output of label predictor is a label vector, whose dimension equals to the number of total content labels, which can be figured out from the result of Formula ~\Ref{clustering_kmeans}. The label predictor block shares the network structure and weights for predicting the behavior labels of adjacent nodes of positive edges. The label predictor function is depicted as Formula \Ref{model_mlp_lp}.
\begin{equation}
Bl(v_i| \theta^{Bl}) \triangleq b_i = MLP(h_i | \theta^{Bl}) = Relu(\mathbf{W_5} h_i)  \label{model_mlp_lp} \text{.}
\end{equation}

\subsubsection{Multi-task learning}
For adjacent nodes of any given positive edge, their behavior labels is supposed to be the same and as close as possible to the content label of one node. For this objective, contrastive learning loss is introduced and depicted as Formula \Ref{model_labelloss}.
\begin{equation}
\begin{split}
L_c(\theta) = \min_{\theta} { \sum\limits_{e^+ \in E, v_i, v_j \in E^+}{CE(c_i, b_i \circ b_j ) + 
\frac{1}{2} \sum\limits_{i = 1,2,3,5}{\lambda_i\Vert\mathbf{W_i}\Vert^2} } }\label{model_labelloss} 
\text{,}
\end{split}
\end{equation}
\noindent where $CE(\cdot)$ represents Cross Entropy loss, $\circ$ represents Hadamard product, and $\lambda_i$ is the corresponding weight hyper-parameter.

Finally, the objective loss function is figured out as the weighted sum of pairwise objective loss $L_p$ and contrastive objective loss $L_c$, which is shown as Formula \Ref{model_finalloss}
\begin{equation}
L(\theta) = \min_{\theta} \sum\limits_{e^+ \in E, e^- \in E', v_i, v_j \in E^+}{\gamma L_h + (1 - \gamma) L_c } \text{,}
\label{model_finalloss}
\end{equation}
\noindent where $\gamma$ is a weight hyper-parameter, which is range from 0 to 1.

\section{Experiments}  \label{section_exp}
Below we evaluate the performance of our algorithm PSG on OGB dataset ogbl-ddi and ogbl-collab ~\cite{hu2020ogb}.

\subsection{Datasets and task} As mentioned before, the ogbl-ddi dataset is a homogeneous, unweighted, undirected graph, representing the drug-drug interaction network~\cite{wishart2018drugbank}, containing 4267 nodes and 2135822 edges. Similarly, the ogbl-collab dataset is also an undirected graph, representing a subset of the collaboration between authors indexed by MAG~\cite{wang2020microsoft}. In addition, all nodes have 128-dimensional content embedding associated with published papers, and two attributes: the year and the edge weight, representing the number of co-authored papers published in that year. The task is to predict links given information on already known links for the two given datasets, which can be reformed as to rank true links higher than false links.

\subsection{Baselines} We compare our algorithm PSG with PLNLP~\cite{wang2021pairwise}, GraphSAGE+Edge Attr~\cite{shitao2021struct}, GIDN/AGDN~\cite{wang2022gidn, sun2020adaptive} and other GNN based algorithms on the leaderboad of OGB. PLNLP introduces pairwise loss function and data augmentation based on the framework of GraphSAGE node encoder and MLP link predictor. GraphSAGE+Edge Attr utilizes distance encoding to integrate edge features based on GraphSAGE encoder and MLP link predictor. GIDN/AGDN further utilizes hop-wise attention within k-hop neighborhood sub-graph to improve message aggregation of sub-graph features. 

\subsection{Metrics and settings} Recommended by OGB, the evaluation metric is Hits@K, which is the ratio of positive edges that are ranked at K-place or above, among a set of approximately 100,000 randomly-sampled negative edges. Here, $K = 20$ for ogbl-ddi and $K=50$ for ogbl-collab. Commonly, each result is achieved by running 10 times with batch size 65536, Adam optimizer, and learning rate is 0.001, activation function is relu, and dropout is 0.3. For other parameters, it is summarized in Tabel ~\ref{tab_param}. The experiments are executed in PyG with Pytorch under the circumstance of Tesla V100 GPU(32G RAM).

\begin{table}[htbp]
    \centering
    \caption{Hyperparameters of PSG model.}\label{tab_param}
    \begin{tabular}{lccc}
    \hline
    Hyperparameter                                    & DDI & COLLAB         \\ \hline
    Epoches                                & 500     &    1500           \\
    Parameters                                & 3499009       &   60823091          \\
    Number of shared GNN layers                                & 2     &      1         \\
    Number of readout MLP layers                                & 2         &     2      \\
    Number of edge features                              & 3    &        131        \\
    Node embedding dimensions       & 512     &    256         \\
    Hidden channels of GNN\&MLP    & 512   &    256            \\
    Number of Label MLP layers                                & None       &     2      \\
    Number of Node Content Clustering                 & None       &     50      \\
    val as input                                & None       &     Yes      \\
    year for filtering training samples                    & None       &     2011      \\
    $\gamma$                                & None       &     0.5      \\
    \hline
    \end{tabular}
\end{table}

\subsection{Abalation Study} As a contrast, some important results of baselines are extracted from the ogbl-ddi and ogbl-collab leaderboads and put in the upper parts of Table ~\ref{tab_ddi_result}  and ~\ref{tab_collab_result}. In the lower parts of the two tables, the results are figured out by conducting the listed algorithms in the same environment with different associated settings, which are proposed by corresponding authors. Firstly, in Table ~\ref{tab_ddi_result}, PSG beats the state-of-the-art algorithm PLNLP on the leaderboad until submission and achieves  2.2\% more performance improvement in terms of average test Hits@20. It indicates that edge featuring part is efficiently integrated into the backbone PLNLP network for dataset ogbl-ddi. Secondly, in Table ~\ref{tab_collab_result}, PSGv2 with label task achieves 0.77\% performance gain with respect to PSG and GIDN+PSGv2 also achieves 0.60\% performance gain with respect to GIDN in the same environment. It shows that our contrastive learning part can improve the graph representation learning of GNN networks, and hence can improve the prediction effectiveness of link prediction based GNN models for dataset ogbl-collab. Finally, PSGv2 beats all other GNN models based on our runs and environment. It verifies the effectiveness of PSG network model especially on dataset ogbl-ddi and ogbl-collab.

\begin{table}[htbp]
    \centering
    \caption{Performance of GNN models on dataset ogbl-ddi.}\label{tab_ddi_result}
    \begin{tabular}{lccc}
    \hline
    Algorithm                                    & Test Hits@20  & Val Hits@20             \\ \hline
    SEAL                                & 0.3056 ± 0.0386          & 0.2849 ± 0.0269         \\
    GCN       & 0.3707 ± 0.0507         & 0.5550 ± 0.0208          \\
    GraphSAGE & 0.5390 ± 0.0474         & 0.6262 ± 0.0037        \\
    GCN+JKNet    & 0.6056 ± 0.0869          & 0.6776 ± 0.0095          \\
    GraphSAGE + Edge Attr   & 0.8781 ± 0.0474 & 0.8044 ± 0.0404 \\
    PLNLP   & 0.9088 ± 0.0313 & 0.8242 ± 0.0253 \\  \hline
    PSG(epochs = 400)   & \textbf{0.9118 ± 0.0235} & 0.8161 ± 0.0232 \\
    PSG(epochs = 500)   & \textbf{0.9284 ± 0.0047} & \textbf{0.8306 ± 0.0134} \\
    PSG(epochs = 600)   & \textbf{0.9189 ± 0.0298} & \textbf{0.8562 ± 0.0181} \\ \hline
    \end{tabular}
\end{table}

\begin{table}[htbp]
    \centering
    \caption{Performance of GNN models on dataset ogbl-collab.}\label{tab_collab_result}
    \begin{tabular}{lccc}
    \hline
    Algorithm            & Test Hits@50  & Val Hits@50             \\ \hline
    GCN       & 0.4475 ± 0.0107         & 0.5263 ± 0.0115          \\
    GraphSAGE & 0.4810 ± 0.0081         & 0.5688 ± 0.0077        \\
    GraphSAGE (val as input) & 0.5463 ± 0.0112         & 0.5688 ± 0.0077        \\
    DeeperGCN   & 0.5273 ± 0.0047 & 0.6187 ± 0.0045 \\
    SEAL-nofeat (val as input)        & 0.6474 ± 0.0043          & 0.6495 ± 0.0043        \\
    PLNLP (val as input)   & 0.6872 ± 0.0052 & 1.0000 ± 0.0000 \\  \hline
    PLNLP + SIGN   & 0.7062 ± 0.0023 & 1.0000 ± 0.0000 \\
    GIDN + PSGv2   & \textbf{0.7069 ± 0.0052} & \textbf{0.9632 ± 0.0002} \\
    GIDN   & 0.7027 ± 0.0046 & 0.9621 ± 0.0003 \\
    PSGv2(Label task)   & \textbf{0.6565 ± 0.0064} & \textbf{0.9428 ± 0.0115} \\ 
    PSG(val as input)   & \textbf{0.6505 ± 0.0055} & \textbf{0.9472 ± 0.0085} \\ \hline
    \end{tabular}
\end{table}

\section{Conclusion}  \label{section_conclusion}
In this paper, we propose a PSG algorithm for link prediction tasks, which includes edge featuring, node encoder, link predictor,  pairwise objective loss, label predictor and contrastive learning framework. PSG achieves top 1 performance on ogbl-ddi until submission and top 3 performance on ogbl-collab. The experimental results verify the superiority of PSG. We will optimize contrastive learning, clustering and attention mechanisms in future work.

\bibliographystyle{IEEEtran}
\bibliography{mybibliography}

\begin{thebibliography}{10}
\providecommand{\url}[1]{#1}
\csname url@samestyle\endcsname
\providecommand{\newblock}{\relax}
\providecommand{\bibinfo}[2]{#2}
\providecommand{\BIBentrySTDinterwordspacing}{\spaceskip=0pt\relax}
\providecommand{\BIBentryALTinterwordstretchfactor}{4}
\providecommand{\BIBentryALTinterwordspacing}{\spaceskip=\fontdimen2\font plus
\BIBentryALTinterwordstretchfactor\fontdimen3\font minus
  \fontdimen4\font\relax}
\providecommand{\BIBforeignlanguage}[2]{{%
\expandafter\ifx\csname l@#1\endcsname\relax
\typeout{** WARNING: IEEEtran.bst: No hyphenation pattern has been}%
\typeout{** loaded for the language `#1'. Using the pattern for}%
\typeout{** the default language instead.}%
\else
\language=\csname l@#1\endcsname
\fi
#2}}
\providecommand{\BIBdecl}{\relax}
\BIBdecl

\bibitem{bennett2007netflix}
J.~Bennett, S.~Lanning \emph{et~al.}, ``The netflix prize,'' in
  \emph{Proceedings of KDD cup and workshop}, vol. 2007.\hskip 1em plus 0.5em
  minus 0.4em\relax Citeseer, 2007, p.~35.

\bibitem{adamic2003friends}
L.~A. Adamic and E.~Adar, ``Friends and neighbors on the web,'' \emph{Social
  networks}, vol.~25, no.~3, pp. 211--230, 2003.

\bibitem{zhao2018bipartite}
Q.~Zhao, H.~Yu, Z.~Ming, H.~Hu, G.~Ren, and H.~Liu, ``The bipartite network
  projection-recommended algorithm for predicting long non-coding rna-protein
  interactions,'' \emph{Molecular Therapy-Nucleic Acids}, vol.~13, pp.
  464--471, 2018.

\bibitem{stanfield2017drug}
Z.~Stanfield, M.~Co{\c{s}}kun, and M.~Koyut{\"u}rk, ``Drug response prediction
  as a link prediction problem,'' \emph{Scientific reports}, vol.~7, no.~1, pp.
  1--13, 2017.

\bibitem{nickel2015review}
M.~Nickel, K.~Murphy, V.~Tresp, and E.~Gabrilovich, ``A review of relational
  machine learning for knowledge graphs,'' \emph{Proceedings of the IEEE}, vol.
  104, no.~1, pp. 11--33, 2015.

\bibitem{wu2020comprehensive}
Z.~Wu, S.~Pan, F.~Chen, G.~Long, C.~Zhang, and S.~Y. Philip, ``A comprehensive
  survey on graph neural networks,'' \emph{IEEE transactions on neural networks
  and learning systems}, vol.~32, no.~1, pp. 4--24, 2020.

\bibitem{zhou2020graph}
J.~Zhou, G.~Cui, S.~Hu, Z.~Zhang, C.~Yang, Z.~Liu, L.~Wang, C.~Li, and M.~Sun,
  ``Graph neural networks: A review of methods and applications,'' \emph{AI
  Open}, vol.~1, pp. 57--81, 2020.

\bibitem{hu2020ogb}
W.~Hu, M.~Fey, M.~Zitnik, Y.~Dong, H.~Ren, B.~Liu, M.~Catasta, and J.~Leskovec,
  ``Open graph benchmark: Datasets for machine learning on graphs,''
  \emph{arXiv preprint arXiv:2005.00687}, 2020.

\bibitem{wishart2018drugbank}
D.~S. Wishart, Y.~D. Feunang, A.~C. Guo, E.~J. Lo, A.~Marcu, J.~R. Grant,
  T.~Sajed, D.~Johnson, C.~Li, Z.~Sayeeda \emph{et~al.}, ``Drugbank 5.0: a
  major update to the drugbank database for 2018,'' \emph{Nucleic acids
  research}, vol.~46, no.~D1, pp. D1074--D1082, 2018.

\bibitem{li2020distance}
P.~Li, Y.~Wang, H.~Wang, and J.~Leskovec, ``Distance encoding: Design provably
  more powerful neural networks for graph representation learning,''
  \emph{Advances in Neural Information Processing Systems}, vol.~33, pp.
  4465--4478, 2020.

\bibitem{hamilton2017inductive}
W.~Hamilton, Z.~Ying, and J.~Leskovec, ``Inductive representation learning on
  large graphs,'' \emph{Advances in neural information processing systems},
  vol.~30, 2017.

\bibitem{wang2021pairwise}
Z.~Wang, Y.~Zhou, L.~Hong, Y.~Zou, and H.~Su, ``Pairwise learning for neural
  link prediction,'' \emph{arXiv preprint arXiv:2112.02936}, 2021.

\bibitem{wang2022gidn}
Z.~Wang, Y.~Guo, J.~Zhao, Y.~Zhang, H.~Yu, X.~Liao, H.~Jin, B.~Wang, and T.~Yu,
  ``Gidn: A lightweight graph inception diffusion network for high-efficient
  link prediction,'' \emph{arXiv preprint arXiv:2210.01301}, 2022.

\bibitem{liben2007link}
D.~Liben-Nowell and J.~Kleinberg, ``The link-prediction problem for social
  networks,'' \emph{Journal of the American society for information science and
  technology}, vol.~58, no.~7, pp. 1019--1031, 2007.

\bibitem{jeh2002simrank}
G.~Jeh and J.~Widom, ``Simrank: a measure of structural-context similarity,''
  in \emph{Proceedings of the eighth ACM SIGKDD international conference on
  Knowledge discovery and data mining}, 2002, pp. 538--543.

\bibitem{grover2016node2vec}
A.~Grover and J.~Leskovec, ``node2vec: Scalable feature learning for
  networks,'' in \emph{Proceedings of the 22nd ACM SIGKDD international
  conference on Knowledge discovery and data mining}, 2016, pp. 855--864.

\bibitem{wang2017predictive}
Z.~Wang, C.~Chen, and W.~Li, ``Predictive network representation learning for
  link prediction,'' in \emph{Proceedings of the 40th international ACM SIGIR
  conference on research and development in information retrieval}, 2017, pp.
  969--972.

\bibitem{kipf2016semi}
T.~N. Kipf and M.~Welling, ``Semi-supervised classification with graph
  convolutional networks,'' \emph{arXiv preprint arXiv:1609.02907}, 2016.

\bibitem{zhang2018link}
M.~Zhang and Y.~Chen, ``Link prediction based on graph neural networks,''
  \emph{Advances in neural information processing systems}, vol.~31, 2018.

\bibitem{liu2021swin}
Z.~Liu, Y.~Lin, Y.~Cao, H.~Hu, Y.~Wei, Z.~Zhang, S.~Lin, and B.~Guo, ``Swin
  transformer: Hierarchical vision transformer using shifted windows,'' in
  \emph{Proceedings of the IEEE/CVF International Conference on Computer
  Vision}, 2021, pp. 10\,012--10\,022.

\bibitem{snyder2018x}
D.~Snyder, D.~Garcia-Romero, G.~Sell, D.~Povey, and S.~Khudanpur, ``X-vectors:
  Robust dnn embeddings for speaker recognition,'' in \emph{2018 IEEE
  international conference on acoustics, speech and signal processing
  (ICASSP)}.\hskip 1em plus 0.5em minus 0.4em\relax IEEE, 2018, pp. 5329--5333.

\bibitem{devlin2018bert}
J.~Devlin, M.-W. Chang, K.~Lee, and K.~Toutanova, ``Bert: Pre-training of deep
  bidirectional transformers for language understanding,'' \emph{arXiv preprint
  arXiv:1810.04805}, 2018.

\bibitem{defferrard2016convolutional}
M.~Defferrard, X.~Bresson, and P.~Vandergheynst, ``Convolutional neural
  networks on graphs with fast localized spectral filtering,'' \emph{Advances
  in neural information processing systems}, vol.~29, 2016.

\bibitem{levie2018cayleynets}
R.~Levie, F.~Monti, X.~Bresson, and M.~M. Bronstein, ``Cayleynets: Graph
  convolutional neural networks with complex rational spectral filters,''
  \emph{IEEE Transactions on Signal Processing}, vol.~67, no.~1, pp. 97--109,
  2018.

\bibitem{huang2022graph}
C.~Huang, M.~Li, F.~Cao, H.~Fujita, Z.~Li, and X.~Wu, ``Are graph convolutional
  networks with random weights feasible?'' \emph{IEEE Transactions on Pattern
  Analysis and Machine Intelligence}, 2022.

\bibitem{gilmer2017neural}
J.~Gilmer, S.~S. Schoenholz, P.~F. Riley, O.~Vinyals, and G.~E. Dahl, ``Neural
  message passing for quantum chemistry,'' in \emph{International conference on
  machine learning}.\hskip 1em plus 0.5em minus 0.4em\relax PMLR, 2017, pp.
  1263--1272.

\bibitem{velivckovic2017graph}
P.~Veli{\v{c}}kovi{\'c}, G.~Cucurull, A.~Casanova, A.~Romero, P.~Lio, and
  Y.~Bengio, ``Graph attention networks,'' \emph{arXiv preprint
  arXiv:1710.10903}, 2017.

\bibitem{lin2022structure}
X.~Lin, Z.~Li, P.~Zhang, L.~Liu, C.~Zhou, B.~Wang, and Z.~Tian,
  ``Structure-aware prototypical neural process for few-shot graph
  classification,'' \emph{IEEE Transactions on Neural Networks and Learning
  Systems}, 2022.

\bibitem{liu2021anomaly}
Y.~Liu, Z.~Li, S.~Pan, C.~Gong, C.~Zhou, and G.~Karypis, ``Anomaly detection on
  attributed networks via contrastive self-supervised learning,'' \emph{IEEE
  transactions on neural networks and learning systems}, vol.~33, no.~6, pp.
  2378--2392, 2021.

\bibitem{gao2015consistency}
W.~Gao and Z.-H. Zhou, ``On the consistency of auc pairwise optimization,'' in
  \emph{Twenty-Fourth International Joint Conference on Artificial
  Intelligence}, 2015.

\bibitem{shitao2021struct}
S.~Lu and J.~Yang, ``Link prediction with structural information,''
  \url{https://github.com/lustoo/OGB_link_prediction/blob/main/Link%20prediction%20with%20structural%20information.pdf},
  2021.

\bibitem{murtagh1991multilayer}
F.~Murtagh, ``Multilayer perceptrons for classification and regression,''
  \emph{Neurocomputing}, vol.~2, no. 5-6, pp. 183--197, 1991.

\bibitem{sculley2010web}
D.~Sculley, ``Web-scale k-means clustering,'' in \emph{Proceedings of the 19th
  international conference on World wide web}, 2010, pp. 1177--1178.

\bibitem{wang2020microsoft}
K.~Wang, Z.~Shen, C.~Huang, C.-H. Wu, Y.~Dong, and A.~Kanakia, ``Microsoft
  academic graph: When experts are not enough,'' \emph{Quantitative Science
  Studies}, vol.~1, no.~1, pp. 396--413, 2020.

\bibitem{sun2020adaptive}
C.~Sun and G.~Wu, ``Adaptive graph diffusion networks with hop-wise
  attention,'' \emph{arXiv preprint arXiv:2012.15024}, 2020.

\end{thebibliography}
\end{document}